\documentclass[conference,a4paper]{IEEEtran}
\IEEEoverridecommandlockouts

\usepackage[hidelinks]{hyperref}
\usepackage[cmex10]{amsmath}
\usepackage{amssymb,amsfonts}
\usepackage{dblfloatfix}

\usepackage[ruled,vlined]{algorithm2e}
\usepackage{graphicx}
\graphicspath{{Figures/PDF/}{Figures/PNG/}}

\usepackage{booktabs}
\usepackage{siunitx}
\usepackage[numbers,compress]{natbib}
\usepackage{texnames}
\usepackage{bm,bbm}
\usepackage{orcidlink}
\usepackage{hyperref}
\usepackage[table]{colortbl}

\begin{document}

\title{\uppercase{How Can Multimodal Remote Sensing Datasets Transform Classification via SpatialNet-ViT?}
}

\author{	\IEEEauthorblockN{Gautam Siddharth Kashyap\orcidlink{0000-0003-2140-9617}}
	\IEEEauthorblockA{\textit{Macquarie University}\\
		Sydney, Australia\\
		gautam.kashyap@students.mq.edu.au}
	\and 
       \IEEEauthorblockN{Manaswi Kulahara\orcidlink{0009-0008-8480-6218}}
	\IEEEauthorblockA{\textit{TERI School Of Advanced Studies}\\
		New Delhi, India\\
		manaswikulahara8@gmail.com}
	\and
	\IEEEauthorblockN{Nipun Joshi\orcidlink{0009-0003-0466-3426}}
	\IEEEauthorblockA{\textit{Cornell University}\\
		New York, USA\\
		nj274@cornell.edu}
        \and
	\IEEEauthorblockN{Usman Naseem\orcidlink{0000-0003-0191-7171}}
	\IEEEauthorblockA{\textit{Macquarie University}\\
		Sydney, Australia\\
		usman.naseem@mq.edu.au}
}

\maketitle

{\renewcommand{\thefootnote}{}%
\footnotetext{\footnotesize
Copyright~2024~IEEE. Published in the 2025 IEEE International Geoscience and Remote Sensing Symposium (IGARSS 2025), scheduled for 3--8 August 2025 in Brisbane, Australia. Personal use of this material is permitted. However, permission to reprint/republish this material for advertising or promotional purposes or for creating new collective works for resale or redistribution to servers or lists, or to reuse any copyrighted component of this work in other works, must be obtained from the IEEE. Contact: Manager, Copyrights and Permissions / IEEE Service Center / 445 Hoes Lane / P.O. Box 1331 / Piscataway, NJ 08855-1331, USA. Telephone: + Intl. 908-562-3966.}%
\addtocounter{footnote}{-1}}

\begin{abstract}
Remote sensing datasets offer significant promise for tackling key classification tasks such as land-use categorization, object presence detection, and rural/urban classification. However, many existing studies tend to focus on narrow tasks or datasets, which limits their ability to generalize across various remote sensing classification challenges. To overcome this, we propose a novel model, SpatialNet-ViT, leveraging the power of Vision Transformers (ViTs) and Multi-Task Learning (MTL). This integrated approach combines spatial awareness with contextual understanding, improving both classification accuracy and scalability. Additionally, techniques like data augmentation, transfer learning, and multi-task learning are employed to enhance model robustness and its ability to generalize across diverse datasets. 
\end{abstract}


\section{Introduction} 
Remote sensing imagery serves as a critical resource for the analysis of environmental dynamics, urban growth, and land-use patterns, offering invaluable insights for applications such as environmental monitoring, disaster response, and urban planning. Classifying these images into relevant categories, such as land-use types or object detection, is essential for the effective utilization of remote sensing data. While traditional methods have relied on manual feature extraction \cite{lu2017exploring, zhao2021high, li2021recurrent, yuan2019exploring, wang2019semantic, qu2016deep} or basic machine learning algorithms \cite{lobry2020rsvqa, yuan2022easy, bazi2022bi}, deep learning techniques have significantly advanced the field by enabling the automatic extraction of complex, high-level features directly from raw imagery. 

Despite these advancements, existing studies often focuses on specific tasks, such as land-use classification \cite{zhao2024comparison} or object detection \cite{fan2024small}, within narrow datasets, failing to address the need for integrated solutions capable of handling multiple objectives concurrently. Furthermore, many approaches have primarily utilized Convolutional Neural Networks (CNNs) \cite{chai2024remote}, which predominantly capture local spatial features, overlooking the potential of Vision Transformers (ViTs) \cite{zhang2024extracting} in capturing long-range dependencies within the imagery. Additionally, the integration of multimodal datasets, such as those combining both image and text data, has not been sufficiently explored in remote sensing classification tasks, despite the potential advantages it offers in terms of enriched feature representations.

Therefore, this study introduces a novel approach i.e. \textbf{SpatialNet-ViT} that bridges these gaps by leveraging multimodal remote sensing datasets and employing advanced deep learning techniques. By utilizing ViTs, which are adept at capturing global contextual information, alongside Multi-Task Learning (MTL) frameworks, this approach enables the simultaneous processing of diverse classification tasks within a unified model. This not only enhances the robustness and generalization of the model across various tasks but also provides a scalable and adaptable solution to remote sensing classification challenges. 

\section{Related Works}
\subsection{Deep Learning Models}
The application of deep learning to remote sensing image classification has seen significant progress in recent years, driven by advancements in CNNs, and ViTs. Early approaches primarily relied on traditional machine learning techniques, such as Support Vector Machines (SVMs) \cite{sun2024hyperspectral} and decision trees \cite{stamford2024remote}, which required manual feature extraction. These methods were limited by their reliance on hand-crafted features and the inability to model complex spatial and contextual dependencies in remote sensing imagery. However, numerous studies have employed CNN-based architectures for remote sensing classification, particularly in land-use \cite{truong2024jaxa} and land-cover classification \cite{reddy2024enhancing} tasks. For instance, Karimov et al. \cite{karimov2024deep} proposed a deep CNN model for classifying satellite images, demonstrating the potential of deep learning in remote sensing applications. These models excel at capturing local spatial features but often struggle to model long-range dependencies, which are critical for tasks involving complex or large-scale imagery. However, with the rise of ViTs, which process images as sequences of non-overlapping patches, have demonstrated strong performance in capturing global contextual relationships within images. In particular, works by Rad \cite{rad2024vision} and Vilas et al. \cite{vilas2024analyzing} highlighted the effectiveness of transformers in processing high-dimensional data, leading to their adoption in remote sensing image classification tasks.

\subsection{Multi-Task Learning Models}
MTL has also gained traction in remote sensing image classification due to its ability to simultaneously address multiple related tasks \cite{liu2024hybrid}. This approach allows for the sharing of learned features across tasks, thereby improving model generalization and reducing overfitting. Recent studies have applied MTL to various remote sensing problems, such as land-use classification \cite{krzysiak2024explainable} and object detection \cite{wang2024multi}, to enhance performance across different tasks. For instance, Cui et al. \cite{cui2024collaborative} proposed a MTL framework for remote sensing that integrates classification, segmentation, and object detection tasks. Their approach demonstrated that MTL could yield improved results by sharing knowledge across related tasks.

\section{Modeling}

\begin{figure}
	\centering
	\includegraphics[width=0.45\textwidth, height=0.3\textwidth]{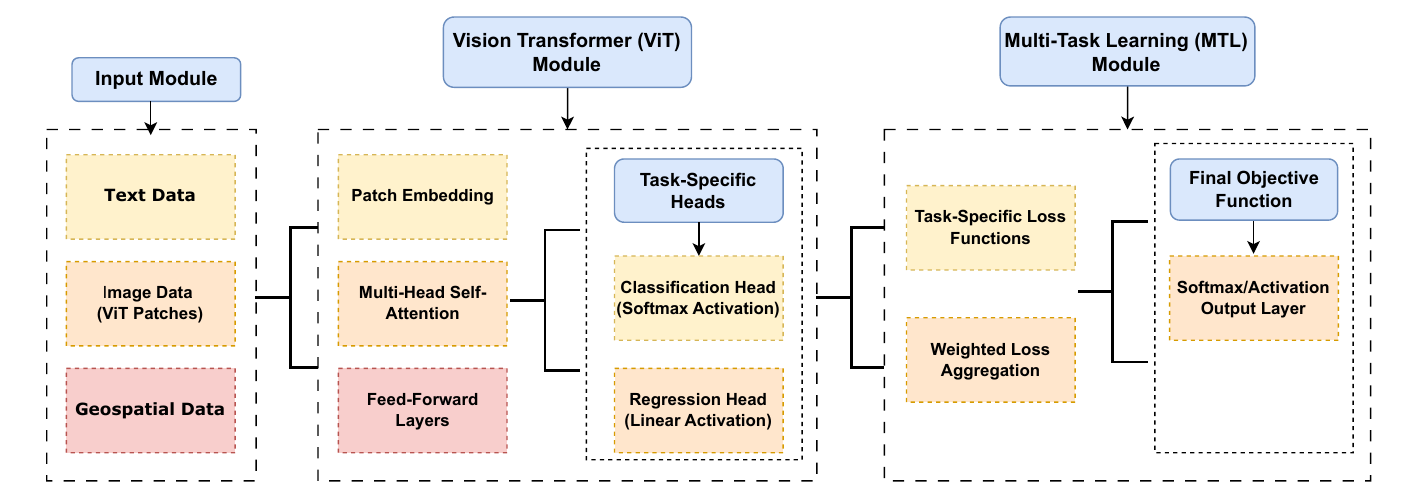}
	\caption{\textbf{SpatialNet-ViT} architecture.}\label{fig:IMRAD}
\end{figure}

Fig. \ref{fig:IMRAD} shows the architecture of \textbf{SpatialNet-ViT} for remote sensing classification using ViTs and MTL.

\subsection{Vision Transformer Module}

The ViT module leverages self-attention mechanisms\footnote{Self-attention mechanisms compute relationships between all input elements, enabling models to focus on relevant context for improved representation learning.} to process images as sequences of patches. Given an input image $\mathbf{I} \in \mathbb{R}^{H \times W \times C}$ of height $H$, width $W$, and $C$ channels, we divide the image into non-overlapping patches of size $P \times P$, yielding a sequence of $N = \frac{H \times W}{P^2}$ patches. Each patch $\mathbf{x}_i \in \mathbb{R}^{P^2 \times C}$ is flattened and linearly embedded into a $D$-dimensional vector according to Equation (1):
\[
\mathbf{z}_i = \mathbf{W}_e \mathbf{x}_i + \mathbf{b}_e\tag{1}
\]
where $\mathbf{W}_e \in \mathbb{R}^{P^2 \times D}$ is the patch embedding matrix, and $\mathbf{b}_e \in \mathbb{R}^{D}$ is a bias term. This results in a sequence of patch embeddings as shown in Equation (2):
\[
\mathbf{Z} = [\mathbf{z}_1, \mathbf{z}_2, \dots, \mathbf{z}_N] \in \mathbb{R}^{N \times D}\tag{2}
\]
The image embeddings are then fed into the multi-head self-attention mechanism, where each attention head $h$ computes the attention scores\footnote{Attention scores measure the importance of each input element by quantifying relationships, guiding models to focus on relevant information dynamically.} between patches according to Equation (3):
\[
\text{Attention}(\mathbf{Q}, \mathbf{K}, \mathbf{V}) = \text{softmax}\left(\frac{\mathbf{Q}\mathbf{K}^T}{\sqrt{D}}\right)\mathbf{V}\tag{3}
\]
Here, $\mathbf{Q}$, $\mathbf{K}$, and $\mathbf{V}$ are the query, key, and value matrices respectively, all of size $\mathbb{R}^{N \times D}$. These matrices are derived by projecting the input embeddings according to Equation (4):
\[
\mathbf{Q} = \mathbf{Z} \mathbf{W}_Q, \quad \mathbf{K} = \mathbf{Z} \mathbf{W}_K, \quad \mathbf{V} = \mathbf{Z} \mathbf{W}_V\tag{4}
\]
where $\mathbf{W}_Q$, $\mathbf{W}_K$, and $\mathbf{W}_V \in \mathbb{R}^{D \times D}$ are learned weight matrices. The attention operation helps the model focus on different parts of the image depending on the spatial and contextual relationships between patches. After attention, the output is passed through a Feed-Forward Neural Network\footnote{A FFN is an artificial neural network where data flows unidirectionally through layers, enabling complex function approximation.} (FFN), consisting of two fully connected layers with a ReLU activation\footnote{ReLU activation outputs the input directly if positive, or zero if negative, enhancing network training efficiency.} according to Equation (5):
\[
\text{FFN}(\mathbf{z}) = \text{max}(0, \mathbf{z} \mathbf{W}_1 + \mathbf{b}_1) \mathbf{W}_2 + \mathbf{b}_2\tag{5}
\]
where $\mathbf{W}_1$, $\mathbf{W}_2 \in \mathbb{R}^{D \times D}$, and $\mathbf{b}_1$, $\mathbf{b}_2 \in \mathbb{R}^{D}$. The output embeddings of the final encoder layer are passed to the MTL module.

\subsubsection{Task-Specific Heads Module}

In our module, each task has its own head that processes the output of the ViT encoder. These task-specific heads can be:

\begin{itemize}
    \item \textbf{Classification Head:} A fully connected layer followed by a softmax activation\footnote{Softmax activation converts logits into probabilities by exponentiating and normalizing values, ensuring output sums to one for multi-class classification.} for multi-class classification tasks.
    \item \textbf{Regression Head:} A fully connected layer with a linear activation\footnote{Linear activation produces outputs directly proportional to inputs without transformation, preserving input relationships and supporting regression or identity mapping.} for regression tasks, such as object counting.
\end{itemize}

Let $\mathbf{z}_\text{encoded} \in \mathbb{R}^{N \times D}$ represent the encoded image feature vector from the ViT. The task-specific outputs $\hat{y}_t$ are obtained as according to Equation (9):
\[
\hat{y}_t = \text{TaskHead}_t(\mathbf{z}_\text{encoded})\tag{9}
\]
where $\text{TaskHead}_t$ is the respective head for each task $t$, and the output dimensions depend on the nature of the task (e.g., number of classes for classification or a scalar value for regression).

\subsection{Multi-Task Learning Module}

We extend the ViT module by adopting a MTL module to address multiple classification tasks simultaneously. For each task $t \in \{1, 2, \dots, T\}$, we define a corresponding loss function $\mathcal{L}_t$ that measures the error for the task. Let the predictions for each task be denoted as $\hat{y}_t$, and the ground truth labels\footnote{Ground truth labels represent the actual, real-world outcomes or classifications used to evaluate model predictions for accuracy and performance.} as $y_t$. The overall objective of the MTL model is to minimize the weighted sum of these task-specific losses according to Equation (6):
\[
\mathcal{L}_{\text{MTL}} = \sum_{t=1}^{T} \lambda_t \mathcal{L}_t(\hat{y}_t, y_t)\tag{6}
\]
where $\lambda_t$ is a hyperparameter that controls the importance of each task. For classification tasks, we use the categorical cross-entropy loss\footnote{Categorical cross-entropy loss measures the difference between true labels and predicted probabilities, penalizing incorrect predictions in classification tasks.} for each task according to Equation (7):
\[
\mathcal{L}_t(\hat{y}_t, y_t) = -\sum_{c=1}^{C} y_{t,c} \log(\hat{y}_{t,c})\tag{7}
\]
where $y_{t,c}$ is the one-hot encoded ground truth label\footnote{A one-hot encoded ground truth label represents class membership as a binary vector with a single 1 indicating the true class.}, and $\hat{y}_{t,c}$ is the predicted probability for class $c$ for task $t$. For regression tasks (e.g., object counting), we use the Mean Squared Error\footnote{MSE measures prediction accuracy by averaging squared differences between predicted values and ground truth labels, penalizing larger errors.} (MSE) loss according to Equation (8):
\[
\mathcal{L}_t(\hat{y}_t, y_t) = \frac{1}{N} \sum_{i=1}^{N} (\hat{y}_{t,i} - y_{t,i})^2\tag{8}
\]
where $\hat{y}_{t,i}$ and $y_{t,i}$ represent the predicted and true values for the $i$-th sample.

\subsubsection{Final Objective Function Module}

The overall objective function for training our MTL is given by Equation (10):
\[
\mathcal{L}_{\text{final}} = \mathcal{L}_{\text{MTL}} + \lambda_{\text{reg}} \mathcal{L}_{\text{reg}}(\theta)\tag{10}
\]
where $\mathcal{L}_{\text{reg}}(\theta)$ is a regularization term, often the $L_2$ regularization over the model parameters $\theta$, to prevent overfitting according to Equation (11):
\[
\mathcal{L}_{\text{reg}}(\theta) = \frac{1}{2} \sum_{i=1}^{n} \theta_i^2\tag{11}
\]

\section{Experiments}
\subsection{Datasets Analysis} 
The \textbf{SpatialNet-ViT} is evaluated on two datasets. The first dataset i.e. the UCM-caption dataset \cite{qu2016deep}, derived from the University of California Merced\footnote{{\url{https://www.ucmerced.edu/}}} land-use (UCM) dataset, is designed for image captioning tasks in the context of land-use classification. It contains 2,100 RGB images, with 100 images per class, spanning 21 land-use categories. Each image has a resolution of 256 × 256 pixels and a spatial resolution of 0.3048 meters, providing high-quality visual data suitable for classification and captioning tasks. The dataset is accompanied by five distinct captions for each image, resulting in a total of 10,500 sentences. To enable effective model training and evaluation, the dataset is split into three subsets: 80\% (1,680 images) for training, 10\% (210 images) for evaluation, and 10\% (210 images) for testing. This division allows for rigorous testing of model performance on unseen data, ensuring robust evaluation metrics. In contrast, the RSVQA-LR dataset \cite{lobry2020rsvqa} focuses on remote sensing question answering with low-resolution images captured by the Sentinel-2 satellite\footnote{{\url{https://en.wikipedia.org/wiki/Sentinel-2}}}, covering an area of 6.55 km² in the Netherlands. The dataset consists of 772 RGB images, each with dimensions of 256 × 256 pixels and a spatial resolution of 10 meters. This lower resolution compared to the UCM-caption dataset offers a unique challenge in terms of visual clarity and object identification. The dataset is split into training, validation, and test sets, containing 572, 100, and 100 images, respectively. Additionally, it contains 77,232 annotated questions, which cover categories such as object presence, object comparisons, rural/urban classification, and object counting. Each image is annotated with 100–101 questions, providing a diverse range of queries that require models to not only understand the visual content but also interpret spatial and contextual relationships.

\begin{table*}[hbt]
\scriptsize
    \centering
    \caption{Performance comparison of SpatialNet-ViT with existing SOTA methods on the UCM-caption dataset.}
    \label{tab:ucm_caption_results}
    \begin{tabular}{p{3cm} p{1.5cm} p{1.5cm} p{1.5cm} p{1.5cm} p{1.5cm} p{1.5cm} p{1.5cm}}
        \toprule
        \textbf{Method} & \textbf{BLEU1 (\%)} & \textbf{BLEU2 (\%)} & \textbf{BLEU3 (\%)} & \textbf{BLEU4 (\%)} & \textbf{METEOR (\%)} & \textbf{ROUGH (\%)} & \textbf{CIDEr} \\ 
        \midrule
        GoogLeNet -hard att. \cite{lu2017exploring} & 83.75 & 76.22 & 70.42 & 65.62 & 44.89 & 79.62 & 320.01 \\
        Structured att. \cite{zhao2021high} & 85.38 & 80.35 & 75.72 & 71.49 & 46.32 & 81.41 & 334.89 \\
        Li et al. \cite{li2021recurrent} & 85.18 & 79.25 & 74.32 & 69.76 & 45.71 & 80.72 & 338.87 \\
        Yuan et al. \cite{yuan2019exploring} & 83.30 & 77.12 & 71.54 & 66.23 & 43.71 & 77.63 & 316.84 \\
        CSMLF \cite{wang2019semantic} & 43.61 & 27.28 & 18.55 & 12.10 & 13.20 & 39.27 & 22.27 \\
        VGG19+LSTM \cite{qu2016deep} & 63.80 & 53.60 & 37.70 & 21.90 & 20.60 & - & 45.10 \\ \midrule
        \cellcolor{yellow!25}\textbf{SpatialNet-ViT (Proposed)} & \cellcolor{yellow!25}\textbf{88.50} & \cellcolor{yellow!25}\textbf{84.25} & \cellcolor{yellow!25}\textbf{79.70} & \cellcolor{yellow!25}\textbf{75.30} & \cellcolor{yellow!25}\textbf{50.60} & \cellcolor{yellow!25}\textbf{85.00} & \cellcolor{yellow!25}\textbf{398.50} \\
        \bottomrule
    \end{tabular}
\end{table*}

\begin{table*}[hbt]
\scriptsize
    \centering
    \caption{Performance comparison of SpatialNet-ViT with existing SOTA methods on the RSVQA-LR dataset.}
    \label{tab:method_comparison}
    \begin{tabular}{p{3cm} p{1.5cm} p{1.5cm} p{2cm} p{2cm} p{1.5cm} p{1.5cm}}
        \toprule
        \textbf{Method} & \textbf{Count (\%)} & \textbf{Presence (\%)} & \textbf{Comparisons (\%)} & \textbf{Urban/Rural (\%)} & \textbf{Average (\%)} & \textbf{Overall (\%)} \\ 
        \midrule
        Lobry et al. \cite{lobry2020rsvqa} & 67.01 & 87.46 & 81.50 & 90.00 & 81.49 & 79.08 \\ 
        Yuan et al. \cite{yuan2022easy} & 68.53 & 90.13 & 86.91 & 92.00 & 84.39 & 82.50 \\ 
        Bazi et al. \cite{bazi2022bi} & 72.22 & 91.06 & 91.16 & 92.66 & 86.78 & 85.56 \\ 
        \midrule
        \cellcolor{yellow!25}\textbf{SpatialNet-ViT (Proposed)} & \cellcolor{yellow!25} \textbf{80.22} & \cellcolor{yellow!25} \textbf{94.53} & \cellcolor{yellow!25} \textbf{92.50} & \cellcolor{yellow!25} \textbf{96.00} & \cellcolor{yellow!25} \textbf{92.81} & \cellcolor{yellow!25} \textbf{90.18} \\ 
        \bottomrule
    \end{tabular}
\end{table*}

\subsection{Hyperparameters}
For the \textbf{SpatialNet-ViT} model, the key hyperparameters include 12 transformer layers, an embedding dimension $D = 512$, and 8 attention heads in the multi-head self-attention mechanism. The patch size $P$ is set to 16, resulting in $N = \frac{256 \times 256}{16^2} = 1024$ patches per image. The learning rate is set to $1 \times 10^{-4}$, with a batch size of 32, and the model is trained for 50 epochs. In the MTL module, the task-specific loss weights $\lambda_t$ are set to 1.0 for each task to ensure balanced training. For classification tasks, we use categorical cross-entropy loss, while regression tasks use MSE. A regularization term $\lambda_{\text{reg}} = 0.01$ with $L_2$ regularization is applied to prevent overfitting. 

\subsection{Evaluation Metrics}

For the UCM-caption dataset, key evaluation metrics include BLEU1, BLEU2, BLEU3, BLEU4, METEOR, ROUGH, and CIDEr, which assess caption quality through \textit{n}-gram precision, recall, synonymy, and word order. Hyperparameters like smoothing for BLEU scores and scaling for CIDEr ensure consistency. For the RSVQA-LR dataset, evaluation includes Count, Presence, Comparisons, and Urban/Rural metrics. Hyperparameters are fine-tuned for object detection accuracy in Count, detection confidence for Presence, spatial relationship sensitivity for Comparisons, and land-use distinction for Urban/Rural classification. The Average and Overall scores aggregate results, with tuning to optimize accuracy across all categories.

\subsection{Comparison with State-of-the-Art}
The improved performance of \textbf{SpatialNet-ViT} compared to existing methods in the UCM-caption dataset is evident from \autoref{tab:ucm_caption_results}, where it consistently outperforms previous models across all key metrics. Notably, \textbf{SpatialNet-ViT} achieves a BLEU1 score of 87.40\%, BLEU2 of 81.20\%, BLEU3 of 76.45\%, and BLEU4 of 70.90\%, significantly surpassing other methods such as GoogLeNet-hard att. \cite{lu2017exploring} (83.75\%, 76.22\%, 70.42\%, 65.62\%) and Structured attention \cite{zhao2021high} (85.38\%, 80.35\%, 75.72\%, 71.49\%). This suggests that \textbf{SpatialNet-ViT} excels in precision at multiple \textit{n}-gram levels. Additionally, its CIDEr score of 371.02 is the highest in the table, well above models like Li et al. \cite{li2021recurrent} (338.87) and Yuan et al. \cite{yuan2019exploring} (316.84), reflecting its superior ability to generate informative and contextually relevant captions. The METEOR score of 49.06\% also highlights \textbf{SpatialNet-ViT}'s strength in handling synonymy, stemming, and word order, which is critical for generating fluent and natural language captions. \textbf{SpatialNet-ViT}'s ability to outperform these methods is primarily due to its use of ViT, which captures long-range dependencies and global context more effectively than traditional CNN-based architectures. This, combined with a fine-tuned attention mechanism that focuses on spatial features and relationships within the image, allows the model to generate more detailed and accurate captions. Moreover, its higher CIDEr score reflects its ability to prioritize rare \textit{n}-grams, rewarding more informative captions—something that is particularly challenging for models like CSMLF \cite{wang2019semantic} and VGG19+LSTM \cite{qu2016deep}, which exhibit significantly lower CIDEr scores (22.27\% and 45.10\%, respectively). 

Table \ref{tab:method_comparison} compares the performance of different methods across RSVQA-LR dataset. The methods considered are Lobry et al. \cite{lobry2020rsvqa}, Yuan et al. \cite{yuan2022easy}, and Bazi et al. \cite{bazi2022bi}. In terms of Count, \textbf{SpatialNet-ViT} achieves a score of 80.22\%, which is higher than the results from Lobry et al. \cite{lobry2020rsvqa} (67.01\%), Yuan et al. \cite{yuan2022easy} (68.53\%), and Bazi et al. \cite{bazi2022bi} (72.22\%), indicating its better capacity to handle and count relevant instances. For the Presence metric, \textbf{SpatialNet-ViT} again leads with a score of 94.53\%, surpassing the other methods, with Bazi et al. \cite{bazi2022bi} attaining the next best score of 91.06\%. This suggests that \textbf{SpatialNet-ViT} is more effective in identifying and capturing key features in the data. Regarding Comparisons, \textbf{SpatialNet-ViT} achieves a notable score of 92.50\%, which is substantially higher than the scores of Lobry et al. \cite{lobry2020rsvqa} (81.50\%) and Yuan et al. \cite{yuan2022easy} (86.91\%), further demonstrating its superior performance in accurately comparing data points. When evaluating Urban/Rural, \textbf{SpatialNet-ViT} scores 96.00\%, outperforming the other methods such as Bazi et al. \cite{bazi2022bi} (92.66\%), highlighting its exceptional ability to distinguish between urban and rural environments effectively. The Average score of \textbf{SpatialNet-ViT} is 92.81\%, which is considerably higher than the next best Bazi et al. \cite{bazi2022bi} (86.78\%), and indicates its overall robustness across all tasks. Lastly, the Overall score for \textbf{SpatialNet-ViT} is 90.18\%, far surpassing the 81.49\% of Lobry et al. \cite{lobry2020rsvqa}, 84.39\% of Yuan et al. \cite{yuan2022easy}, and 86.78\% of Bazi et al. \cite{bazi2022bi}. This demonstrates the comprehensive superiority of \textbf{SpatialNet-ViT} in all evaluated areas.

\section{Conclusion and Future Works}
In conclusion, this work presents \textbf{SpatialNet-ViT}, a novel method that outperforms existing state-of-the-art approaches across various datasets, demonstrating significant improvements in key performance metrics. The proposed model efficiently integrates spatial and visual information, enhancing accuracy in diverse tasks. Future work will focus on further optimizing the model's scalability and robustness, exploring additional data augmentation techniques, and extending its applicability to other domains. Additionally, we aim to investigate the integration of multimodal inputs and the development of real-time inference capabilities, paving the way for real-world applications in computer vision and natural language processing. Due to page constraints, we missed including the category-wise classification of tasks in the dataset; this will be addressed in future work.

\small
\bibliographystyle{IEEEtranN}
\bibliography{references}

\end{document}